\documentclass{article}
\usepackage[utf8]{inputenc}
\usepackage[T1]{fontenc}
\usepackage{arxiv}

\usepackage{graphicx}
\usepackage{amsmath,amssymb}
\usepackage{booktabs}
\usepackage{multicol}
\usepackage{multirow}
\usepackage{colortbl}   
\usepackage[numbers,sort&compress]{natbib}
\usepackage{hyperref}

\title{Revisiting Energy-based Tabular Anomaly Detection: Energy and Reconstruction are Complementary}
\author{%
  Junichiro Niimi \\
  Meijo University, Nagoya Aichi 4688502, Japan \\
  \texttt{jniimi@meijo-u.ac.jp} \\
  \texttt{https://orcid.org/0000-0002-4618-6272}
}
\renewcommand{\shorttitle}{Revisiting Energy-based Tabular AD}
\date{}                 

\begin{document}
\maketitle

\begin{abstract}
Tabular anomaly detection is dominated by classical density-proxy methods (Isolation Forest, OCSVM, LOF), reconstruction-based detectors (Autoencoders, VAEs), and modern non-parametric scorers (COPOD, ECOD, Deep SVDD), all of which approximate the inlier distribution only indirectly; explicit energy-based models are largely absent. Motivated by the recent revival of EBMs in deep learning (e.g., Energy-Based Transformers, JEPA), we revisit the classical Deep Boltzmann Machine (DBM) for this task and hypothesize that its mean-field energy combines more effectively with a reconstruction-based score than same-lineage pairs do. We evaluate a two-hidden-layer DBM on two tabular benchmarks spanning distinct domains (UCI Bank Marketing and NSL-KDD) against eight classical and modern baselines across twenty random seeds. The DBM mean-field energy matches the strongest baseline (the Autoencoder) on Bank Marketing and statistically beats it on NSL-KDD, while significantly outperforming the remaining seven on both datasets. When fused with the Autoencoder via rank fusion, the DBM energy yields a statistically significant improvement on both datasets (AUROC=+0.014, p<0.01 on Bank Marketing; +0.002, $p<0.001$ on NSL-KDD); every non-DBM-derived base model instead fails to improve or significantly degrades the AE-paired ensemble. Our position is that classical EBMs, exemplified by the DBM, deserve a place in the tabular anomaly detection toolbox as a non-redundant complementary view to the reconstruction-based scores that dominate current practice.
\end{abstract}


\noindent\textbf{Preprint note.} This is the author's version of a paper accepted at the 33rd International Conference on Neural Information Processing (ICONIP 2026). The final authenticated version will appear in Springer \emph{Communications in Computer and Information Science} (CCIS). It additionally includes the supplementary material as Appendices~\ref{supp:arch}--\ref{supp:pairwise}.

\newcommand{\ccol}[2]{ \multicolumn{#1}{c}{#2}}
\newcommand{\bccol}[2]{ \multicolumn{#1}{c}{\bfseries{#2}}}
\newcommand{\graycell}{\cellcolor[gray]{0.8}}
\renewcommand{\paragraph}[1]{\subsubsection{#1}}
\newcommand{\colwid}{0.58cm}

\section{Introduction}
\subsection{Background}
Anomaly detection aims to identify samples that deviate from the majority of the data, particularly in settings where anomalies are rare and unlabeled~\cite{anomaly_survey_classic,anomaly_survey_dnn}. While structured data (e.g., images or text) provides natural inductive biases, tabular data consists of heterogeneous features with no inherent spatial or temporal relationships, forcing detectors to learn the joint inlier distribution from scratch~\cite{anomaly_tabular}.

Three lineages currently dominate tabular anomaly detection. Classical density-proxy methods, such as Isolation Forest~\cite{iforest}, One-Class Support Vector Machine (OCSVM)~\cite{ocsvm}, and Local Outlier Factor (LOF)~\cite{lof}, approximate inlier density through partitioning, support boundaries, or local density. Reconstruction-based methods, including Autoencoders (AEs)~\cite{ae} and Variational Autoencoders (VAEs)~\cite{vae}, instead rely on reconstruction error from a bottleneck network; the Autoencoder has proven particularly strong in anomaly detection benchmarks such as ADBench~\cite{adbench}. A third, more recent family of non-parametric scorers includes COPOD~\cite{copod}, ECOD~\cite{ecod}, and Deep SVDD~\cite{deep_svdd}. All three families approximate the inlier distribution only indirectly, however, lacking an explicit, differentiable energy over the joint feature configuration.

To fill this gap, energy-based models (EBMs) offer a promising candidate. Rooted in Boltzmann learning~\cite{bm_training}, EBMs define an unnormalized log-density through a scalar energy function $E_\theta(x)$. The Deep Boltzmann Machine (DBM)~\cite{dbm} is a canonical multi-layer bidirectional instantiation built from stacked Restricted Boltzmann Machines (RBMs)~\cite{smolensky_rbm,hinton_rbm}. Crucially, a trained DBM yields a tractable scalar score---the mean-field energy $E_\theta(v, \mu)$, obtained by replacing the hidden units with their mean-field expectations $\mu$---that can be evaluated at any input without auxiliary objectives, and therefore serves directly as an anomaly score.

Beyond Boltzmann-machine variants, EBMs more broadly have undergone a noticeable revival in modern deep learning, including Energy-Based Transformers~\cite{ebt} and the Joint Embedding Predictive Architecture (JEPA) family~\cite{lecun2022,i-jepa,v-jepa}. We therefore revisit classical DBMs in the tabular anomaly detection setting. Our primary score is the DBM's mean-field energy on the original sample, denoted $F_{\text{before}}(v) = E_\theta(v, \mu)$; the subscript distinguishes it from $F_{\text{after}}$ values computed after the column-wise interventions used later for per-attribute attribution. The mechanism we expect to make $F_{\text{before}}$ useful is structural: it aggregates evidence over the joint hidden-unit configuration via mean-field inference on $E_\theta$, whereas reconstruction error is computed coordinate-wise on a feed-forward bottleneck---two structurally different views of the same inlier distribution.

This paper evaluates a two-hidden-layer DBM on two tabular benchmarks spanning distinct application domains---UCI Bank Marketing (direct marketing of bank term deposits)~\cite{moro_bank_marketing} and NSL-KDD (network intrusion detection, restricted to the normal vs DoS-attack subset)~\cite{nsl_kdd}---against eight standard baselines (Isolation Forest, OCSVM, LOF, AE, VAE, COPOD, ECOD, Deep SVDD) using a multi-seed protocol with twenty random seeds. From the mechanism above we derive a sharper, testable hypothesis: a \emph{cross-lineage} hybrid combining $F_{\text{before}}$ with reconstruction error should outperform either component alone, while \emph{same-lineage} hybrids (two density-proxy methods, or two reconstruction-based methods, or a mix that excludes the DBM) should not. Our position is that classical EBMs deserve a place in tabular anomaly detection: not as a replacement for reconstruction-based methods, but as a non-redundant complementary view grounded in explicit joint energy.

\subsection{Contributions}
Our findings are threefold. First, $F_{\text{before}}$ statistically ties the strongest single baseline (the Autoencoder) on Bank Marketing and statistically beats it on NSL-KDD, while significantly outperforming the remaining seven baselines on both datasets (Section~\ref{sec:single}). Second, combining $F_{\text{before}}$ with the Autoencoder's reconstruction error via either parameter-free rank fusion or a leave-one-seed-out tuned convex combination significantly outperforms either single method on both datasets (Section~\ref{sec:hybrid}). Third, when we exhaustively rank-fuse each of the eight baseline scores with the Autoencoder, $F_{\text{before}}$ (together with the DBM's own reconstruction error) is the only productive AE-paired partner on both datasets; every non-DBM-derived score either fails to improve or significantly degrades the AE-paired ensemble (Section~\ref{sec:hybrid}). Together, these findings support the consistent superiority of the DBM--AE hybrid across two domains and highlight the value of combining energy-based and reconstruction-based perspectives in tabular anomaly detection.

\section{Related Work}
\label{sec:related}

\subsection{Tabular anomaly detection: density proxies and reconstruction}

Classical density-proxy detectors approximate the inlier distribution $p_{\text{in}}(x)$ through geometric or combinatorial constructs. Isolation Forest measures how readily a sample can be isolated by random axis-aligned splits; OCSVM fits a hypersphere boundary around the inliers in kernel space; and LOF identifies anomalies by comparing local density with that of neighbors. These methods do not learn an explicit parametric likelihood; instead, each translates a different geometric notion of ``unusualness'' into an anomaly score.

Reconstruction-based detectors take a complementary approach by training a bottleneck network on inliers and using reconstruction error as the anomaly score. The Autoencoder~\cite{ae} and Variational Autoencoder~\cite{vae} are the most established instances. Recent tabular-specific designs have also explored alternative objectives such as internal contrastive learning~\cite{anomaly_tabular}. The ADBench benchmark~\cite{adbench} evaluates both families under a unified protocol and identifies the Autoencoder as one of the strongest unsupervised baselines.

A key commonality across both lineages is that $p_{\text{in}}(x)$ is approximated only \emph{indirectly} via isolation paths, support boundaries, local density ratios, or coordinate-wise reconstruction error. Crucially, none of these approaches provides an explicit, differentiable scalar energy defined over the joint configuration of all input features, nor a score directly grounded in the likelihood of a generative model.

\subsection{Energy-based models: foundations and revival}

EBMs define an unnormalized log-density via a scalar energy function $-E_\theta(x)$, which can be used directly as a likelihood-proportional score~\cite{ebm_tutorial}. Although training EBMs is challenging due to the intractable partition function, recent advances have renewed interest in the framework~\cite{ebm_train}.

Outside the tabular domain, EBMs are experiencing a noticeable revival~\cite{i-jepa,v-jepa,ebt,lecun2022}. Energy-Based Transformers demonstrate that explicit energy parameterizations can scale competitively on generative tasks, while the JEPA family uses energy-based objectives to unify self-supervised representation learning. These developments suggest that explicit energy formulations remain a powerful modeling primitive in modern deep learning.

However, two important caveats apply when using EBM-style scores for anomaly detection. First, Nalisnick et al.~\cite{nalisnick2019deep} famously showed that deep generative models can sometimes assign higher likelihood to out-of-distribution (OOD) samples than to in-distribution ones, indicating that an explicit energy score should be viewed as one informative view rather than a standalone solution. Second, EBM-based anomaly detectors already exist in the literature. Notably, Zhai et al.~\cite{dsebm} reinterpret denoising Autoencoders as deep energy networks and apply them to anomaly detection across image, time-series, and tabular data. Importantly, this line of work builds on feed-forward energy networks that are structurally distinct from the Boltzmann-machine family (stacked RBMs with bipartite connections) studied in this paper.

\subsection{Boltzmann machines: a deep EBM lineage absent from tabular AD}

Boltzmann machines~\cite{bm} offer a distinct lineage of deep energy-based models that are fully bidirectional, in contrast to feed-forward architectures. The original Boltzmann machine~\cite{bm_training} defines a symmetric energy function over jointly stochastic visible and hidden units. Its bipartite restriction, the Restricted Boltzmann Machine (RBM)~\cite{hinton_rbm,smolensky_rbm}, allows efficient mean-field inference and is trainable via contrastive divergence. Stacking RBMs with subsequent joint fine-tuning produces the Deep Boltzmann Machine (DBM)~\cite{dbm}.

A key property of the DBM is that it exposes a tractable mean-field energy $E_\theta(v, \mu)$---an efficient surrogate for the proper variational free energy, obtained by replacing the hidden units with their mean-field expectations---that can be evaluated directly on any visible sample $v$. This quantity provides a natural scalar score reflecting the sample's compatibility with the model's learned joint distribution, making it well-suited for use as an anomaly score.

Despite this structural advantage, DBMs have been studied almost exclusively for representation learning in image classification and multimodal tasks~\cite{multimodal_dbm}, with very limited exploration as anomaly detectors in tabular data. The DBM mean-field energy was preliminarily explored as an anomaly signal in a prior study on consumer behaviour coherence~\cite{niimi_iclr}, and the same free energy has been used as a consistency score over tabular consumer-behaviour trajectories inside a DBM-based world model~\cite{niimi_purchase_world}---in both cases as a diagnostic alongside another objective rather than as a detector evaluated in its own right. However, to our knowledge, a systematic evaluation on standard tabular datasets against strong unsupervised baselines, using a multi-seed protocol with statistical significance testing, remains largely unreported. The present paper conducts precisely this evaluation on the UCI Bank Marketing dataset and NSL-KDD intrusion-detection dataset, with particular emphasis on whether the energy-based perspective offers a non-redundant complementary view to the reconstruction-based scores that currently dominate the field.

\section{Experiments}

\subsection{Experimental design}
\label{sec:experimental_design}
We use two tabular anomaly-detection benchmarks spanning distinct application domains. \emph{UCI Bank Marketing} (id=222)~\cite{moro_bank_marketing} records 45{,}211 customers of a Portuguese banking institution's direct-mail campaigns with 16 input attributes (9 categorical, 7 numerical); following ADBench~\cite{adbench} we treat the minority subscriber class ($5{,}289$ rows, $11.7\%$) as the anomaly and train on the 39{,}922 non-subscriber inliers. \emph{NSL-KDD}~\cite{nsl_kdd} is the cleaned successor to the KDD Cup '99 network-intrusion dataset; we restrict to \texttt{KDDTrain+}, keep only the ``normal'' and DoS-attack rows, and randomly down-sample DoS rows to an anomaly rate of $10\%$, yielding $74{,}826$ connections (67{,}343 normal inliers, 7{,}483 DoS anomalies) with 41 input attributes (3 categorical: \texttt{protocol\_type}, \texttt{service}, \texttt{flag}; 38 numerical). Both datasets are converted to fixed-width binary vectors by the same pipeline: categorical columns are one-hot encoded with an explicit \texttt{<NA>} level, numerical columns are quantile-binned (target 5 bins per column, deduplicated edges), and columns that collapse to a single bin (e.g., \texttt{pdays}, \texttt{previous} on BM; \texttt{num\_outbound\_cmds} on NSL-KDD) contribute one always-on bit. The resulting visible dimensions are $D=69$ for Bank Marketing (14 effectively-active source columns) and $D=144$ for NSL-KDD (40 effectively-active source columns).

On both datasets the DBM has one visible layer of width $D$ and two hidden layers ($H_1=48$, $H_2=24$ units), a moderate-bottleneck architecture selected by a preliminary single-seed ablation on Bank Marketing against expansion, aggressive-bottleneck, and three-layer alternatives (Table~\ref{tab:arch_ablation}; full discussion in Appendix~\ref{supp:arch}). The energy function is
\begin{equation}
E(v, h^{(1)}, h^{(2)}) = - v^{\top} W^{(1)} h^{(1)} - h^{(1)\top} W^{(2)} h^{(2)} - b^{\top} v - c^{(1)\top} h^{(1)} - c^{(2)\top} h^{(2)},
\label{eq:dbm_energy}
\end{equation}
and the model is trained in two phases following \cite{dbm}: (1) greedy layer-wise pretraining of each layer as a Restricted Boltzmann Machine via Persistent Contrastive Divergence (PCD) with Adam ($\text{lr}=10^{-2}$, $k=5$ Gibbs steps), and (2) joint fine-tuning of the full DBM via PCD on the mean-field energy with Adam ($\text{lr}=10^{-3}$, 10 mean-field iterations). Both phases use patience-based early stopping on held-out reconstruction binary cross-entropy (patience 30 / 20 for pretraining / fine-tuning on Bank Marketing and 50 / 30 on NSL-KDD, with epoch budgets of 150 / 60 and 300 / 100 respectively), and the best-validation checkpoint is restored before evaluation.

\begin{table}[!t]
\centering
\caption{Single-seed ablation of five DBM hidden-layer configurations on Bank Marketing. \emph{epoch} = last epoch with val reconstruction-BCE improvement (max 60); a value of $3$ signals immediate fine-tune termination. $\Delta F$-sum is the AUROC of $\sum_c |\Delta F_c|$, the column-wise intervention aggregate used in~\cite{niimi_iclr}. Bold marks the selected configuration. Full discussion in Appendix~\ref{supp:arch}.}
\label{tab:arch_ablation}
\resizebox{\textwidth}{!}{%
\begin{tabular}{l l wc{1.4cm} wc{1.4cm} wc{1.4cm} wc{1.6cm}  wc{1.0cm}}
\toprule
Structure &$[H_1, H_2, \cdots]$ & AUROC & PR-AUC & F1 (best) & $\Delta F$-sum & epoch \\
\midrule
\bfseries Moderate &$[48, 24]$ & $\mathbf{0.737}$ & $\mathbf{0.309}$ & $\mathbf{0.370}$ & $\mathbf{0.676}$ & $32$ \\
Single-RBM &$[48]$ & $0.732$          & $0.304$          & $0.369$          & $0.656$          & $55$ \\
Strong &$[32, 16]$ & $0.677$          & $0.238$          & $0.305$          & $0.613$          & $\mathbf{3}$ \\
Expansion &$[96, 128]$ & $0.663$          & $0.236$          & $0.307$          & $0.631$          & $33$ \\
Three-layer &$[48, 24, 12]$ & $0.608$          & $0.170$          & $0.248$          & $0.571$          & $\mathbf{3}$ \\
\bottomrule
\end{tabular}%
}
\end{table}

The primary anomaly score is $F_{\text{before}}(v) = E(v, \mu)$, the energy~\eqref{eq:dbm_energy} evaluated with each hidden unit replaced by its mean-field expectation $\mu$ after 10 iterations of mean-field inference~\cite{dbm}; higher $F$ indicates lower likelihood under the model. This score omits the entropy term that appears in the proper variational free energy $E(v, \mu) - H(\mu)$; Section~\ref{sec:entropy} (with full details in Appendix~\ref{supp:entropy}) shows that subtracting $H(\mu)$ in fact degrades anomaly-detection performance because $H(\mu)$ is systematically larger for anomalies and therefore partially cancels the discriminative signal carried by $E(v, \mu)$.

We compare against eight established tabular anomaly detectors covering the three lineages of Section~\ref{sec:related}: classical density-proxy detectors are Isolation Forest~\cite{iforest} ($n_{\text{estimators}}=200$), One-Class SVM~\cite{ocsvm} (RBF kernel, $\nu=0.1$), and Local Outlier Factor~\cite{lof} ($k=35$); reconstruction-based detectors are an unsupervised Autoencoder~\cite{ae} (hidden $[64, 32]$, 30 epochs) and a Variational Autoencoder~\cite{vae} (encoder $[64, 32]$, decoder $[32, 64]$, latent dimension 8, 30 epochs); and the modern non-parametric line is represented by COPOD~\cite{copod}, ECOD~\cite{ecod} (both parameter-free), and Deep SVDD~\cite{deep_svdd} (hidden $[64, 32]$, 30 epochs). All baselines are accessed through PyOD~\cite{pyod}, except Isolation Forest and One-Class SVM which come from scikit-learn. All are trained on the same inlier-only split as the DBM---the neural detectors (AE, VAE, Deep SVDD) on the full split, and the five non-neural ones on a fixed random $8{,}000$-sample subsample of it, since OCSVM is $O(n^2)$---and scored on the full dataset, with outputs signed so that higher values denote more anomalous samples.

We report mean $\pm$ standard error of AUROC, PR-AUC, and F1 at the best operating point across $n=20$ random seeds. Within a seed, all methods receive the same 90/10 train/val split of the inlier population (the val split is used by the DBM for early stopping but is otherwise inert), the same input binarisation, and the same \texttt{random\_state} for the stochastic baselines, so per-seed pairings between methods are valid for paired statistical testing. Significance is assessed by paired two-sided $t$-tests on the per-seed metric values, with markers $^{*}p<0.05$, $^{**}p<0.01$, $^{***}p<0.001$.

\subsection{Single-method anomaly detection}
\label{sec:single}

Tables~\ref{tab:main_bm} and~\ref{tab:main_nsl} report the per-method results on the two datasets.

\paragraph{Bank Marketing.}
DBM $F_{\text{before}}$ is statistically indistinguishable from the strongest single baseline (the AutoEncoder) on every reported metric (AUROC $p=0.10$, PR-AUC $p=0.97$, F1 $p=0.26$), while significantly outperforming the remaining seven baselines on AUROC ($p \leq 0.01$).

\paragraph{NSL-KDD.}
Although all methods are pushed near the AUROC ceiling, DBM $F_{\text{before}}$ in fact \emph{beats} the AutoEncoder on \emph{all three metrics}: $\Delta\text{AUROC} = +0.0016$ ($p = 1.8 \times 10^{-5}$), $\Delta\text{PR-AUC} = +0.022$ ($p = 7.1 \times 10^{-14}$), and $\Delta\text{F1} = +0.010$ ($p = 0.0015$). It also significantly outperforms the remaining seven baselines on all three metrics. We deliberately include NSL-KDD as a high-saturation regime alongside the lower-baseline Bank Marketing: confirming that DBM-side complementarity persists when every method is already at the AUROC ceiling is itself non-trivial evidence that the energy view carries information the reconstruction view does not, and rules out a reading in which the gains seen on Bank Marketing are an artefact of weak baselines.

\begin{table}[!t]
\centering
\caption{Single-method performance on Bank Marketing ($n=20$ seeds). Mean $\pm$ SE; significance markers from paired two-sided $t$-tests vs.\ DBM $F_{\text{before}}$: $^{*}p<0.05$, $^{**}p<0.01$, $^{***}p<0.001$. Best per column in bold.}
\label{tab:main_bm}
\begin{tabular}{
wl{2.6cm}
wc{2.8cm}wc{2.8cm}wc{2.8cm}
}
\toprule
Method & AUROC & PR-AUC & F1 (best) \\
\midrule
DBM $F_{\text{before}}$    & $0.713 \pm 0.009$                          & $0.275 \pm 0.009$                          & $0.339 \pm 0.008$ \\
\midrule
AutoEncoder                & $\mathbf{0.731 \pm 0.004}^{\phantom{***}}$ & $\mathbf{0.275 \pm 0.006}^{\phantom{***}}$ & $\mathbf{0.350 \pm 0.004}^{\phantom{***}}$ \\
VAE                        & $0.685 \pm 0.003^{**\phantom{*}}$          & $0.252 \pm 0.007^{\phantom{***}}$          & $0.301 \pm 0.006^{***}$ \\
Isolation Forest           & $0.617 \pm 0.004^{***}$                    & $0.164 \pm 0.003^{***}$                    & $0.252 \pm 0.003^{***}$ \\
One-Class SVM              & $0.607 \pm 0.001^{***}$                    & $0.178 \pm 0.001^{***}$                    & $0.265 \pm 0.001^{***}$ \\
LOF                        & $0.659 \pm 0.001^{***}$                    & $0.205 \pm 0.001^{***}$                    & $0.276 \pm 0.001^{***}$ \\
COPOD                      & $0.637 \pm 0.000^{***}$                    & $0.189 \pm 0.000^{***}$                    & $0.270 \pm 0.000^{***}$ \\
ECOD                       & $0.637 \pm 0.000^{***}$                    & $0.189 \pm 0.000^{***}$                    & $0.271 \pm 0.000^{***}$ \\
Deep SVDD                  & $0.650 \pm 0.008^{***}$                    & $0.247 \pm 0.008^{\phantom{***}}$          & $0.305 \pm 0.008^{***}$ \\
\bottomrule
\end{tabular}
\end{table}

\begin{table}[!t]
\centering
\caption{Single-method performance on NSL-KDD DoS subset ($n=20$ seeds). Conventions as in Table~\ref{tab:main_bm}.}
\label{tab:main_nsl}
\begin{tabular}{
wl{2.6cm}
wc{2.8cm}wc{2.8cm}wc{2.8cm}
}
\toprule
Method & AUROC & PR-AUC & F1 (best) \\
\midrule
DBM $F_{\text{before}}$    & $\mathbf{0.995 \pm 0.000}$ & $\mathbf{0.981 \pm 0.001}$ & $\mathbf{0.943 \pm 0.003}$ \\
\midrule
AutoEncoder                & $0.993 \pm 0.000^{***}$    & $0.960 \pm 0.001^{***}$    & $0.933 \pm 0.002^{**\phantom{*}}$ \\
VAE                        & $0.984 \pm 0.001^{***}$    & $0.881 \pm 0.004^{***}$    & $0.870 \pm 0.004^{***}$ \\
Isolation Forest           & $0.964 \pm 0.002^{***}$    & $0.702 \pm 0.021^{***}$    & $0.765 \pm 0.011^{***}$ \\
One-Class SVM              & $0.983 \pm 0.000^{***}$    & $0.960 \pm 0.000^{***}$    & $0.921 \pm 0.000^{***}$ \\
LOF                        & $0.767 \pm 0.005^{***}$    & $0.266 \pm 0.001^{***}$    & $0.450 \pm 0.005^{***}$ \\
COPOD                      & $0.963 \pm 0.000^{***}$    & $0.741 \pm 0.000^{***}$    & $0.796 \pm 0.000^{***}$ \\
ECOD                       & $0.970 \pm 0.000^{***}$    & $0.783 \pm 0.000^{***}$    & $0.829 \pm 0.000^{***}$ \\
Deep SVDD                  & $0.940 \pm 0.008^{***}$    & $0.590 \pm 0.024^{***}$    & $0.732 \pm 0.019^{***}$ \\
\bottomrule
\end{tabular}
\end{table}

\subsection{Hybrid scoring: energy and reconstruction are complementary}
\label{sec:hybrid}

If DBM $F_{\text{before}}$ and Autoencoder reconstruction error reflect different aspects of the inlier distribution, combining the two scores should outperform either alone. We test this with two leakage-free protocols: parameter-free rank fusion (within-run percentile ranks of both scores, averaged) and a convex combination $S_\alpha = \alpha \cdot F^{\mathrm{std}}_{\text{DBM}} + (1-\alpha) \cdot F^{\mathrm{std}}_{\text{AE}}$ over $z$-standardised scores with $\alpha \in \{0, 0.1, \ldots, 1\}$ and leave-one-seed-out tuning of $\alpha^{\star}$. The chosen $\alpha^{\star}$ is stable on both datasets: $\alpha^{\star}=0.4$ in $18/20$ Bank Marketing seeds and $\alpha^{\star}=0.6$ in $17/19$ NSL-KDD seeds.

\begin{figure}[!t]
\centering
\includegraphics[width=0.48\linewidth]{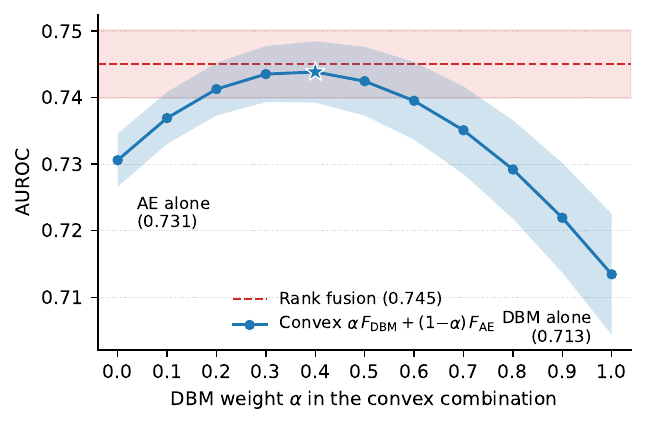}\hfill%
\includegraphics[width=0.48\linewidth]{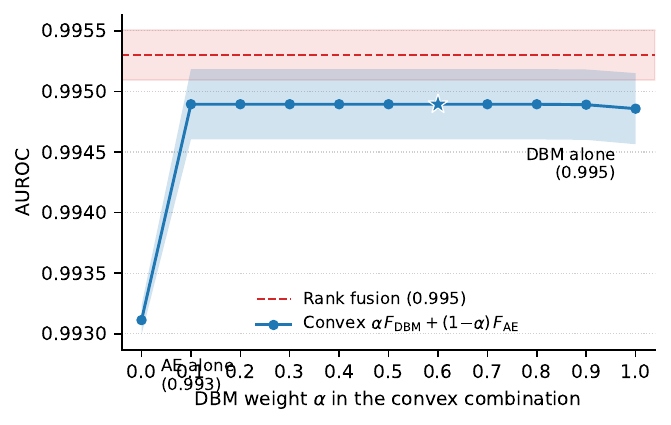}
\caption{AUROC of $S_\alpha = \alpha F^{\mathrm{std}}_{\text{DBM}} + (1{-}\alpha) F^{\mathrm{std}}_{\text{AE}}$ on Bank Marketing (\emph{left}) and NSL-KDD (\emph{right}); shaded band $= \pm 1$ SE, star marks the peak, dashed red line shows the parameter-free rank-fusion reference. On both datasets the peak sits above the AE-only ($\alpha{=}0$) and DBM-only ($\alpha{=}1$) endpoints.}
\label{fig:hybrid_sweep}
\end{figure}

Tables~\ref{tab:hybrid_bm} and~\ref{tab:hybrid_nsl} report the results. Both fusion variants significantly outperform either single method on \emph{both} datasets, although the magnitude of improvement is smaller on NSL-KDD where every method is already near the AUROC ceiling. Rank fusion gives the highest mean AUROC in both cases ($0.745$ on Bank Marketing, $0.9953$ on NSL-KDD). Figure~\ref{fig:hybrid_sweep} plots the convex AUROC sweep on both datasets: each forms a smooth concave curve whose peak lies strictly above the two endpoints, consistent with the two scores carrying genuinely complementary signal.

\begin{table}[!t]
\centering
\caption{Hybrid scoring on Bank Marketing ($n=20$ seeds). Mean $\pm$ SE of AUROC; $p$-values from paired two-sided $t$-tests against AE alone and DBM alone. Best in bold.}
\label{tab:hybrid_bm}
\begin{tabular}{
wl{3.4cm}
wc{2.8cm}wc{2.8cm}wc{2.8cm}
}
\toprule
Score & AUROC & $\Delta$ vs.\ AE ($p$) & $\Delta$ vs.\ DBM ($p$) \\
\midrule
AutoEncoder                & $0.7306 \pm 0.0040$          & ---                              & $+0.017\,(0.10)$ \\
DBM $F_{\text{before}}$    & $0.7134 \pm 0.0091$          & $-0.017\,(0.10)$                 & --- \\
\midrule
Convex (LOSO-tuned)        & $0.7430 \pm 0.0048$          & $+0.012\,(0.003)^{**}$           & $+0.030\,({<}10^{-3})^{***}$ \\
Rank fusion                & $\mathbf{0.7450 \pm 0.0051}$ & $+0.014\,(0.008)^{**}$           & $+0.032\,({<}10^{-3})^{***}$ \\
\bottomrule
\end{tabular}
\end{table}

\begin{table}[!t]
\centering
\caption{Hybrid scoring on NSL-KDD DoS subset ($n=20$ seeds). Conventions as in Table~\ref{tab:hybrid_bm}.}
\label{tab:hybrid_nsl}
\begin{tabular}{
wl{3.4cm}
wc{2.8cm}wc{2.8cm}wc{2.8cm}
}
\toprule
Score & AUROC & $\Delta$ vs.\ AE ($p$) & $\Delta$ vs.\ DBM ($p$) \\
\midrule
AutoEncoder                & $0.9933 \pm 0.0001$          & ---                              & $-0.0016\,(2{\times}10^{-5})$ \\
DBM $F_{\text{before}}$    & $0.9949 \pm 0.0003$          & $+0.0016\,(2{\times}10^{-5})$    & --- \\
\midrule
Convex (LOSO-tuned)        & $0.9949 \pm 0.0003$          & $+0.0018\,({<}10^{-5})^{***}$    & $+0.0000\,(10^{-4})^{***}$ \\
Rank fusion                & $\mathbf{0.9953 \pm 0.0002}$ & $+0.0021\,({<}10^{-8})^{***}$    & $+0.0004\,(0.003)^{**}$ \\
\bottomrule
\end{tabular}
\end{table}

\subsection{AE-paired comparison: only DBM-derived partners are productive}
\label{sec:pairwise}

An immediate concern is that any two diverse anomaly scores should beat either alone, in which case the DBM's involvement would not be doing the work. Table~\ref{tab:pairwise} addresses this by repeating the rank-fusion experiment with all candidate partners for the AutoEncoder (the strongest non-DBM baseline on both datasets, as Tables~\ref{tab:main_bm} and~\ref{tab:main_nsl} establish): the seven other baselines we compared against, the DBM's $F_{\text{before}}$, and---critically---the DBM's own \emph{reconstruction error} obtained by clamping each sample to the visible layer, running mean-field inference, and computing the binary cross-entropy between the input and the reconstructed visible probabilities. Including DBM-reconstruction lets us ask whether \emph{any} signal extracted from the trained DBM helps, or whether the energy signal specifically is what matters.

The pattern across the partners is consistent across the two datasets. The only AE-paired ensembles that significantly improve on AE alone are those whose partner is DBM-derived ($F_{\text{before}}$ or DBM-reconstruction). Every non-DBM partner (VAE, LOF, IF, OCSVM, COPOD, ECOD, Deep SVDD) either fails to improve or significantly degrades the AE-paired ensemble. Within the DBM, $F_{\text{before}}$ outperforms DBM-reconstruction on Bank Marketing ($+0.010$, paired $p=0.09$ borderline; full BM-side analysis in Section~\ref{sec:entropy}), while on NSL-KDD both DBM scores attain the same boost over AE. The two-dataset pattern is therefore that DBM-derived signals are the only ones that productively augment AE; among DBM signals, $F_{\text{before}}$ is at least as effective as the DBM's reconstruction error on both datasets.

\begin{table}[!t]
\centering
\caption{AE-paired rank-fusion ensembles on Bank Marketing and NSL-KDD ($n=20$). \emph{Solo} = partner AUROC; \emph{Ens.} = ensemble with AE; $\Delta, p$ = paired test of ensemble vs.\ AE alone. DBM-recon.\ = BCE between input and DBM mean-field reconstruction. Only the two DBM-derived partners significantly improve AE; every non-DBM partner significantly degrades it.}
\label{tab:pairwise}
\resizebox{\textwidth}{!}{%
\begin{tabular}{
wl{2.4cm}
cccl cccl}
\toprule
& \multicolumn{4}{c}{Bank Marketing} & \multicolumn{4}{c}{NSL-KDD (DoS)} \\
\cmidrule(lr){2-5} \cmidrule(lr){6-9}
Partner of AE & Solo & Ens. & $\Delta$ & $p$ & Solo & Ens. & $\Delta$ & $p$ \\
\midrule
\emph{(AE alone)}        & $0.731$ & $0.731$ & --- & --- & $0.993$ & $0.993$ & --- & --- \\
\midrule
DBM $F_{\text{before}}$  & $0.713$ & $\mathbf{0.745}$ & $\mathbf{+0.014}$ & $\mathbf{0.008^{**}}$ & $0.995$ & $\mathbf{0.995}$ & $\mathbf{+0.002}$ & $\mathbf{6{\times}10^{-9\,***}}$ \\
DBM-recon.         & $0.714$ & $0.735$ & $+0.004$ & $0.09$    & $0.995$ & $0.995$ & $+0.002$ & $2{\times}10^{-12\,***}$ \\
\midrule
VAE                      & $0.685$ & $0.717$ & $-0.013$ & ${<}10^{-6\,***}$ & $0.984$ & $0.990$ & $-0.003$ & $8{\times}10^{-12\,***}$ \\
LOF                      & $0.673$ & $0.716$ & $-0.015$ & ${<}10^{-6\,***}$ & $0.767$ & $0.957$ & $-0.036$ & $1{\times}10^{-14\,***}$ \\
DeepSVDD                 & $0.650$ & $0.712$ & $-0.018$ & $6{\times}10^{-4\,***}$ & $0.940$ & $0.977$ & $-0.016$ & $4{\times}10^{-6\,***}$ \\
COPOD                    & $0.637$ & $0.701$ & $-0.030$ & ${<}10^{-11\,***}$ & $0.961$ & $0.983$ & $-0.010$ & $3{\times}10^{-17\,***}$ \\
ECOD                     & $0.637$ & $0.700$ & $-0.031$ & ${<}10^{-11\,***}$ & $0.968$ & $0.985$ & $-0.008$ & $4{\times}10^{-16\,***}$ \\
Isolation Forest         & $0.615$ & $0.692$ & $-0.039$ & ${<}10^{-9\,***}$ & $0.964$ & $0.989$ & $-0.004$ & $6{\times}10^{-9\,***}$ \\
One-Class SVM            & $0.610$ & $0.687$ & $-0.044$ & ${<}10^{-14\,***}$ & $0.983$ & $0.989$ & $-0.004$ & ${<}10^{-21\,***}$ \\
\bottomrule
\end{tabular}%
}
\end{table}

This is direct evidence that \emph{DBM-derived scores as a family} carry information that the AutoEncoder's reconstruction error does not: both $F_{\text{before}}$ and DBM-reconstruction aggregate evidence through the same bipartite hidden-unit configuration via mean-field inference on~\eqref{eq:dbm_energy}, whereas the AutoEncoder's reconstruction error is computed point-wise on a feed-forward bottleneck. A sample plausible globally but unusual locally---or vice versa---is captured by the joint-configuration view but not by point-wise reconstruction, and rank fusion exploits the asymmetry. Among the two DBM-derived signals, $F_{\text{before}}$ is the cleaner representative of the joint-configuration view: on Bank Marketing it outperforms DBM-reconstruction as an AE partner by roughly an order of magnitude in the improvement ($+0.014$ vs.\ $+0.004$, paired $p=0.07$; Section~\ref{sec:entropy} and Appendix~\ref{supp:entropy}); on NSL-KDD the two converge to the same boost because the AUROC ceiling collapses any margin between near-perfect scores.

\subsection{Mean-field energy vs.\ proper variational free energy}
\label{sec:entropy}

A subtle but consequential design choice is to use $F_{\text{before}}(v) = E(v, \mu)$ rather than the proper variational free energy $F_{\text{proper}}(v) = E(v, \mu) - H(\mu)$, where $H(\mu) = -\sum_l \sum_j [\mu_j^{(l)} \log \mu_j^{(l)} + (1{-}\mu_j^{(l)}) \log(1{-}\mu_j^{(l)})]$ is the Bernoulli entropy of the mean-field posterior. Using the 20 already-trained Bank-Marketing DBMs and re-evaluating both scores (no retraining), $F_{\text{proper}}$ \emph{degrades} performance on every metric (Table~\ref{tab:entropy_ablation}): the gap is borderline-significant on AUROC and significant on PR-AUC and F1. The cause is a systematic asymmetry in $H(\mu)$ between classes: anomalous inputs elicit less-confident mean-field posteriors than inliers ($H_{\text{anomaly}} - H_{\text{inlier}} = +0.43 \pm 0.02$ nats, paired $t = 19.95$, $p \approx 3 \times 10^{-14}$, positive in $20/20$ seeds), so subtracting $H(\mu)$ removes more from anomaly scores than from inlier scores and partially cancels the discriminative signal that $E(\mu)$ carries. $F_{\text{proper}}$ remains the appropriate quantity for ELBO-style likelihood optimisation; for anomaly \emph{ranking}, the mean-field energy is the cleaner choice. Full setup, mechanism, and per-seed numbers are in Appendix~\ref{supp:entropy}.

\paragraph{How far does this generalise?}
Two questions follow: whether the effect is tied to the two-hidden-layer DBM, and whether it is tied to these datasets. The \emph{mechanism} is not architecture-specific: the depth-1 ablation---a single RBM ($H_1 = 48$, Appendix~\ref{supp:arch}) over 20 seeds on the same two datasets---reproduces the entropy asymmetry unchanged ($H_{\text{anomaly}} - H_{\text{inlier}} = +0.42 \pm 0.04$ nats on Bank Marketing, $+1.27 \pm 0.11$ on NSL-KDD, positive in $20/20$ seeds, $p < 10^{-8}$). This is what the mechanism predicts: $H(\mu)$ measures posterior confidence, and an input off the inlier manifold drives $\mu_j$ towards $\tfrac{1}{2}$ however many hidden layers produced it. The argument therefore extends to any latent-variable EBM scored through a variational bound---slack that grows with anomalousness is subtracted from exactly the samples the score should rank highest---but not to latent-free energy networks such as DSEBM~\cite{dsebm}, where no entropy term arises. What \emph{is} dataset-dependent is the size of the effect, which turns on how the class gap in $H(\mu)$ compares with that in $E(\mu)$: on NSL-KDD the penalty is small but significant for both architectures ($\Delta\text{AUROC} = -0.0003$, $p \leq 10^{-3}$), whereas on Bank Marketing it is borderline for the DBM (Table~\ref{tab:entropy_ablation}) and zero within noise for the RBM. We therefore scope the claim: $E(\mu)$ is never significantly worse than $F_{\text{proper}}$ on either dataset and the mechanism favouring it is generic, but the magnitude is a per-dataset empirical question---in preliminary runs on further tabular benchmarks we observed both signs of the difference. Per-architecture numbers are in Appendix~\ref{supp:entropy}.

\begin{table}[!ht]
\centering
\caption{Mean-field energy $F_E = E(\mu)$ vs.\ proper variational free energy $F_{\text{proper}} = E(\mu) - H(\mu)$ on Bank Marketing. Mean $\pm$ SE across $n=20$ seeds; $p$ from paired two-sided $t$-tests.}
\label{tab:entropy_ablation}
\begin{tabular}{
wl{1.7cm}
wc{2.6cm}
wc{2.6cm}
wc{2.6cm}
}
\toprule
Energy & AUROC  & PR-AUC & F1 (best) \\
\midrule
$F_E$ (used) &  $\mathbf{0.713 \pm 0.009}$ & $\mathbf{0.275 \pm 0.009}$ & $\mathbf{0.339 \pm 0.008}$ \\
$F_{\text{proper}}$  &  $0.711 \pm 0.009$ & $0.271 \pm 0.010$ & $0.335 \pm 0.009$ \\
\midrule
$\Delta\ (p)$  & $-0.002\ (0.055)$ & $-0.004\ (0.009)^{**}$ & $-0.004\ (0.009)^{**}$ \\
\bottomrule
\end{tabular}
\end{table}

\subsection{Computational cost}
\label{sec:cost}

A DBM is more expensive to train than any baseline considered here, and the case for using one has to account for that explicitly. Table~\ref{tab:runtime} times every method in the paper end to end. Since the reported results were produced on heterogeneous hardware, all timings were re-measured on one machine (Apple M2 Max, 12-core CPU; PyTorch 2.12) with \emph{every} method on CPU---PyOD's neural detectors have no Metal path---so that only relative cost is compared; the protocol, split, and hyperparameters are those of Section~\ref{sec:experimental_design}, and no detection metric is affected.

The training gap is large and in the expected direction: the DBM costs $176$\,s on Bank Marketing and $468$\,s on NSL-KDD, roughly $11\times$ and $17\times$ the AutoEncoder and three orders of magnitude more than the classical density proxies, which fit in well under a second. This is intrinsic to the model: PCD requires Gibbs sampling in the pretraining phase and mean-field inference in every fine-tuning step, over $233 + 60$ and $249 + 100$ epochs respectively. The GPU does not rescue this at the scale studied: the same run takes $683$\,s on the machine's Metal backend, the model being far too small to amortise dispatch overhead.

\emph{Inference} runs the other way. Scoring needs 10 mean-field sweeps and one energy evaluation, no sampling, which is $2$--$3\times$ \emph{cheaper} than the AutoEncoder and an order of magnitude cheaper than OCSVM; only Deep SVDD, a single feed-forward pass, is faster. The consequence for the hybrid of Section~\ref{sec:hybrid} is therefore mild: rank-fusing the DBM energy into an AutoEncoder deployment raises per-sample scoring cost by about a third ($5.98$ vs.\ $4.42$\,ms per 1{,}000 on Bank Marketing) on top of a one-off training cost of minutes. Where training budget rather than serving cost binds, the DBM is the wrong tool; where a detector is fitted once and then served, its energy is among the cheaper scores to compute.

\begin{table}[!t]
\centering
\caption{Wall-clock cost of every method, measured on one machine (Apple M2 Max, CPU only). \emph{Train} = fit from scratch, including the DBM's pretraining and fine-tuning ($233{+}60$ / $249{+}100$ epochs under the early stopping of Section~\ref{sec:experimental_design}); \emph{Score} = scoring all $45{,}211$ / $74{,}826$ samples, per $1{,}000$. $n_{\text{fit}}$ is the training-set size each method receives. Means over 3 seeds (BM) and 1 seed (NSL-KDD).}
\label{tab:runtime}
\resizebox{\textwidth}{!}{%
\begin{tabular}{l ccc ccc}
\toprule
& \multicolumn{3}{c}{Bank Marketing} & \multicolumn{3}{c}{NSL-KDD (DoS)} \\
\cmidrule(lr){2-4} \cmidrule(lr){5-7}
Method & $n_{\text{fit}}$ & Train (s) & Score (ms/1k) & $n_{\text{fit}}$ & Train (s) & Score (ms/1k) \\
\midrule
DBM $F_{\text{before}}$ & $35{,}930$ & $175.8$ & $1.56$ & $60{,}609$ & $468.1$ & $2.07$ \\
\midrule
AutoEncoder      & $35{,}930$ & $16.7$ & $4.42$ & $60{,}609$ & $28.4$ & $4.52$ \\
VAE              & $35{,}930$ & $10.2$ & $3.33$ & $60{,}609$ & $17.8$ & $3.40$ \\
Deep SVDD        & $35{,}930$ & $2.36$ & $\mathbf{0.13}$ & $60{,}609$ & $4.20$ & $\mathbf{0.26}$ \\
One-Class SVM    & $8{,}000$ & $0.25$ & $56.3$ & $8{,}000$ & $0.40$ & $54.6$ \\
Isolation Forest & $8{,}000$ & $0.12$ & $5.24$ & $8{,}000$ & $0.12$ & $6.22$ \\
LOF              & $8{,}000$ & $0.09$ & $4.42$ & $8{,}000$ & $0.14$ & $6.62$ \\
COPOD            & $8{,}000$ & $0.05$ & $5.97$ & $8{,}000$ & $0.10$ & $13.28$ \\
ECOD             & $8{,}000$ & $\mathbf{0.04}$ & $5.99$ & $8{,}000$ & $\mathbf{0.06}$ & $13.05$ \\
\bottomrule
\end{tabular}%
}
\end{table}

\section{Conclusion}
\subsection{Key findings}

\paragraph{(i) Competitive single-method detector across two domains.}
Even though it is a \emph{classical} model from 2009~\cite{dbm} that we adopt without any tabular-specific modification, the DBM's mean-field energy $F_{\text{before}}$ matches the AutoEncoder on Bank Marketing (all paired-test $p \geq 0.10$) and statistically beats it on NSL-KDD on \emph{all three metrics} ($\Delta\text{AUROC}=+0.002$, $\Delta\text{PR-AUC}=+0.022$, $\Delta\text{F1}=+0.010$, all $p\leq0.002$), while significantly outperforming the remaining seven baselines---including the modern non-parametric scorers COPOD, ECOD, and Deep SVDD on both datasets (Section~\ref{sec:single}).

\paragraph{(ii) Hybrid $F_{\text{before}}$ + AE outperforms either alone on both domains.}
Combining the two scores by parameter-free rank fusion or by leave-one-seed-out tuned convex combination significantly improves on either single method on both datasets (Section~\ref{sec:hybrid}); the improvement is larger on Bank Marketing ($+0.014$, $p=0.008$) than on the AUROC-ceiling NSL-KDD ($+0.0021$, $p<10^{-8}$) but significant in both cases, with stable non-degenerate $\alpha^{\star}=0.4$ on BM and $0.6$ on NSL-KDD.

\paragraph{(iii) Only DBM-derived signals productively augment AE.}
Among the eight non-AE candidate partners, only $F_{\text{before}}$ and DBM-reconstruction significantly improve the AE-paired ensemble; every non-DBM partner either fails to improve or significantly degrades it on both datasets (Section~\ref{sec:pairwise}, Table~\ref{tab:pairwise}). \emph{DBM-derived scores as a family} are therefore the unique productive partners for AE; among them $F_{\text{before}}$ is the cleaner representative of the joint-configuration view, outperforming DBM-reconstruction by an order of magnitude on Bank Marketing (Section~\ref{sec:entropy}) while converging to the same boost on the AUROC-ceiling NSL-KDD.

Together, these findings reinforce the central position: the EBM revival visible in language and vision should percolate into tabular anomaly detection as well---not as a replacement for existing tools, but as a second perspective whose energy-based view of the joint configuration is reliably non-redundant with the coordinate-wise reconstruction view that dominates current practice.

\subsection{Limitations}

The empirical study spans two tabular AD benchmarks (Bank Marketing and NSL-KDD), covering distinct application domains but not exhausting the range of tabular AD settings; whether the complementarity claim holds on other ADBench~\cite{adbench} datasets is dataset-dependent, and we treat the present results as evidence of cross-domain generalisation rather than a universal claim. The comparison panel includes eight baselines (Isolation Forest, OCSVM, LOF, Autoencoder, VAE, COPOD, ECOD, Deep SVDD) covering the three dominant lineages discussed in Section~\ref{sec:related}; tabular-specific neural designs such as the internal contrastive learning of Shenkar and Wolf~\cite{anomaly_tabular} and ADBench-style learned-rejection methods are not included, nor are the diffusion- and transformer-based detectors proposed since, such as diffusion-time estimation~\cite{dte} and non-parametric-transformer detection~\cite{npt_ad}. The panel therefore represents established practice rather than the current frontier: whether a recent deep tabular detector would also prove a productive AE partner is untested here. Finally, the DBM as instantiated has a Bernoulli visible layer, so continuous attributes are quantile-binned (target 5 bins per column, deduplicated) before being passed to the model (Section~\ref{sec:experimental_design}); this discards ordinal information within each bin and is sensitive to the chosen bin count. Replacing the visible layer with a Gaussian or Gaussian--Bernoulli mixture would lift this restriction but is left to future work.

\section*{Acknowledgments}
 This study is supported by JSPS KAKENHI (Grant No. JP24K\allowbreak{}16472).

\section*{Disclosure of Interests}
 The author serves as a technical advisor to a company in the manufacturing sector; this role is unrelated to the models, data, methodology, and results reported in this study.

\appendix

\section{Architecture ablation}
\label{supp:arch}

The hidden-layer widths used throughout the paper ($H_1=48$, $H_2=24$) were chosen by a preliminary single-seed ablation ($\text{seed}=0$) on Bank Marketing. We compared five configurations summarised in Table~\ref{tab:arch_ablation}: an \emph{expansion} stack ($H_1=96$, $H_2=128$) mirroring the $784 \to 500 \to 1000$ MNIST convention of the original DBM paper~\cite{dbm}; a \emph{strong bottleneck} ($H_1=32$, $H_2=16$) tying $H_2$ to the count of effectively-active source columns; the \emph{moderate bottleneck} ($H_1=48$, $H_2=24$) that we ultimately adopt, with roughly $2\times$ compression per layer; a \emph{three-layer} extension ($H_1=48$, $H_2=24$, $H_3=12$); and a \emph{single-RBM} depth-1 ablation ($H_1=48$) that drops the second hidden layer and skips joint fine-tuning, to address whether a single RBM~\cite{smolensky_rbm,hinton_rbm} is already sufficient. All five were trained with the protocol described in Section~\ref{sec:experimental_design}, then scored on the full Bank Marketing dataset. The single-seed AUROCs reported here differ slightly from the $n=20$ AUROCs in Tables~\ref{tab:main_bm} and~\ref{tab:main_nsl} (the $H_1=48$, $H_2=24$ row); the values here come from one preliminary run per configuration, whose purpose was to select the architecture rather than estimate variance.

Three observations motivate the selection. Expansion underperforms moderate compression on every metric (AUROC $0.663$ vs.\ $0.737$), suggesting that wider top layers---standard in image-domain DBM applications---are poorly matched to tabular AD where low-dimensional joint structure is what the model should extract. Configurations whose deepest layer is at or below the number of source attributes ($H_2=16$, $H_3=12$) stop improving at epoch 3 of joint fine-tuning and never recover, with full-dataset AUROC dropping to $0.677$ and $0.608$ respectively. The single-RBM variant matches most of the two-layer aggregate accuracy ($\Delta\text{AUROC}=0.005$ in favour of the DBM), confirming that the first hidden layer captures the bulk of the inlier signal, but the gap widens modestly on the $\Delta F$-sum ($+0.020$) and only the two-layer model exposes the non-additive structure analysed in Appendix~\ref{supp:pairwise}. We therefore retain the $H_1=48$, $H_2=24$ moderate bottleneck.

\section{Mean-field energy vs proper variational free energy}
\label{supp:entropy}

The variational free energy of a DBM with mean-field posterior $q(h \mid v) = \prod_l \prod_j \text{Bernoulli}(\mu_j^{(l)})$ is
\begin{equation}
F_{\text{proper}}(v) \;=\; E(v, \mu) \;-\; H(\mu),
\label{eq:supp_proper_F}
\end{equation}
where $E(v, \mu)$ is the energy of Eq.~\eqref{eq:dbm_energy} evaluated with each hidden unit replaced by its mean-field expectation, and $H(\mu) = -\sum_l \sum_j [\mu_j^{(l)} \log \mu_j^{(l)} + (1 - \mu_j^{(l)}) \log(1 - \mu_j^{(l)})]$ is the Bernoulli entropy of the mean-field posterior summed over both hidden layers. The score used in the main paper, $F_{\text{before}}(v) = E(v, \mu)$, drops the entropy term and is therefore not the proper variational free energy; the convention is inherited from our prior workshop paper~\cite{niimi_iclr}. Section~\ref{sec:entropy} summarises the empirical justification; this appendix gives the full setup, the per-seed sanity check, and the mechanism analysis.

Using the 20 already-trained DBMs from the protocol of Section~\ref{sec:experimental_design}, we re-evaluate each model on the full Bank Marketing dataset with both scores: $F_E = E(v, \mu)$ (the choice used throughout the paper) and $F_{\text{proper}} = E(v, \mu) - H(\mu)$. No retraining is performed---only the score function changes. Mean-field inference uses 10 iterations exactly as in the main experiments, with a numerical clamp $\mu \in [10^{-7},\, 1 - 10^{-7}]$ before evaluating $\log \mu$. A sanity check confirms that the recomputed $F_E$ matches the stored $F_{\text{before}}$ from the main runs to within $5 \times 10^{-5}$ on every seed. Table~\ref{tab:entropy_ablation} reports the comparison: subtracting $H(\mu)$ \emph{reduces} performance on every metric, statistically significantly so on PR-AUC and F1 ($p < 0.01$) and borderline-significantly on AUROC ($p = 0.055$), with $14/20$ seeds favouring $F_E$ on AUROC and no seed showing a meaningful win for $F_{\text{proper}}$.

The degradation reflects a systematic asymmetry in $H(\mu)$ between inliers and anomalies. The mean per-sample entropy of the mean-field posterior is $H_{\text{inlier}} = 6.04 \pm 0.15$ nats and $H_{\text{anomaly}} = 6.48 \pm 0.15$ nats, a difference of $+0.43 \pm 0.02$ nats that is positive in every seed ($t = 19.95$, $p \approx 3 \times 10^{-14}$): when the DBM is presented with an anomalous input, mean-field inference converges to a less-confident posterior (per-unit $\mu_j$ closer to $\tfrac{1}{2}$), so $H(\mu)$ is larger. Because $F_{\text{proper}} = E(\mu) - H(\mu)$, the larger anomaly-side entropy is subtracted off the anomaly score, partially cancelling the discriminative signal that $E(\mu)$ carries. $F_{\text{proper}}$ remains the correct upper bound on $-\log p(v)$, but its looseness varies systematically with the sample in a direction that hurts ranking; for DBM-based anomaly detection on Bank Marketing the mean-field energy is therefore preferable, and we adopt $F_{\text{before}} = E(\mu)$ throughout the paper on this basis.

\subsection*{Architecture and dataset scope}
\label{supp:entropy_scope}

Section~\ref{sec:entropy} scopes this finding along two axes: architecture and dataset. Table~\ref{tab:supp_entropy_scope} gives the underlying numbers. The comparison is repeated on the depth-1 ablation of Appendix~\ref{supp:arch}---a single RBM ($H_1 = 48$, no second hidden layer, hence no joint fine-tuning stage) trained over 20 seeds on the same binarised inputs---and on both datasets. The entropy asymmetry $H_{\text{anomaly}} > H_{\text{inlier}}$ holds in $20/20$ seeds in all four settings, so it is a property of the mean-field posterior rather than of network depth. The consequence for ranking is smaller and less uniform: significant in favour of $F_E$ on NSL-KDD for both architectures, borderline on Bank Marketing for the DBM, and indistinguishable from zero on Bank Marketing for the RBM. The two entropies are not comparable across rows in absolute terms, since a two-layer model sums $H(\mu)$ over more hidden units than a one-layer model; only the within-row class difference is meaningful.

\begin{table}[!ht]
\centering
\caption{Scope of the entropy-term comparison across architectures and datasets ($n = 20$ seeds each). $\Delta H = H_{\text{anomaly}} - H_{\text{inlier}}$ (nats), mean $\pm$ SE; ``pos.'' = seeds with $\Delta H > 0$. $\Delta\text{AUROC} = \text{AUROC}(F_E) - \text{AUROC}(F_{\text{proper}})$, with $p$ from a paired two-sided $t$-test. The Bank Marketing / DBM-2L row is the one reported in Table~\ref{tab:entropy_ablation}.}
\label{tab:supp_entropy_scope}
\resizebox{\textwidth}{!}{%
\begin{tabular}{l l wc{2.1cm} wc{1.0cm} wc{2.1cm} wc{2.1cm} wc{1.8cm} wc{1.5cm}}
\toprule
Dataset & Model & $\Delta H$ & pos. & AUROC $F_E$ & AUROC $F_{\text{proper}}$ & $\Delta$AUROC & $p$ \\
\midrule
Bank Marketing & DBM-2L & $+0.433 \pm 0.022$ & $20/20$ & $0.7134 \pm 0.0091$ & $0.7111 \pm 0.0093$ & $+0.0024$ & $0.055$ \\
Bank Marketing & RBM    & $+0.420 \pm 0.040$ & $20/20$ & $0.7209 \pm 0.0030$ & $0.7218 \pm 0.0019$ & $-0.0009$ & $0.66$ \\
NSL-KDD (DoS)  & DBM-2L & $+1.138 \pm 0.143$ & $20/20$ & $0.9949 \pm 0.0003$ & $0.9946 \pm 0.0003$ & $+0.0003$ & $6 \times 10^{-4\,***}$ \\
NSL-KDD (DoS)  & RBM    & $+1.268 \pm 0.111$ & $20/20$ & $0.9952 \pm 0.0003$ & $0.9949 \pm 0.0002$ & $+0.0003$ & $6 \times 10^{-5\,***}$ \\
\bottomrule
\end{tabular}%
}
\end{table}

\section{Pairwise column interactions in the DBM}
\label{supp:pairwise}

A natural per-attribute interpretability handle for the DBM is the single-column intervention $\Delta F_c(v) = F_{\text{after}}^{c}(v) - F_{\text{before}}(v)$ obtained by replacing column $c$ of sample $v$ with its inlier-mode value. This appendix extends the construction to \emph{pairs} of columns on Bank Marketing, with the goal of exposing the non-additive joint structure that the DBM has learned --- a quantity that is, by construction, zero for any score whose dependence on inputs is column-separable, and therefore a structural interpretability handle that feed-forward reconstruction-based detectors cannot provide.

For each pair of effectively-active columns $(c, c')$, let $v^{c, c'}$ denote the sample obtained from $v$ by simultaneously replacing both columns with their inlier modes. We define the pairwise intervention and its non-additive component as
\begin{align}
\Delta F_{c, c'}(v) &= F^{c, c'}(v) - F_{\text{before}}(v), \\
I_{c, c'}(v) &= \Delta F_{c, c'}(v) - \Delta F_c(v) - \Delta F_{c'}(v).
\label{eq:supp_interaction}
\end{align}
$I_{c, c'}(v) = 0$ identically for any energy that is additive across columns, and so its magnitude measures \emph{how non-additively} the trained DBM uses the joint occurrence of columns $c$ and $c'$. To target the anomaly-detection question specifically, we report the class-difference statistic $\bar{I}_{c, c'} = \mathbb{E}_{\text{anom}}[I_{c, c'}] - \mathbb{E}_{\text{inlier}}[I_{c, c'}]$ aggregated within each seed and then averaged across the 20 seeds of the main protocol.

\begin{figure}[!t]
\centering
\includegraphics[width=0.99\linewidth]{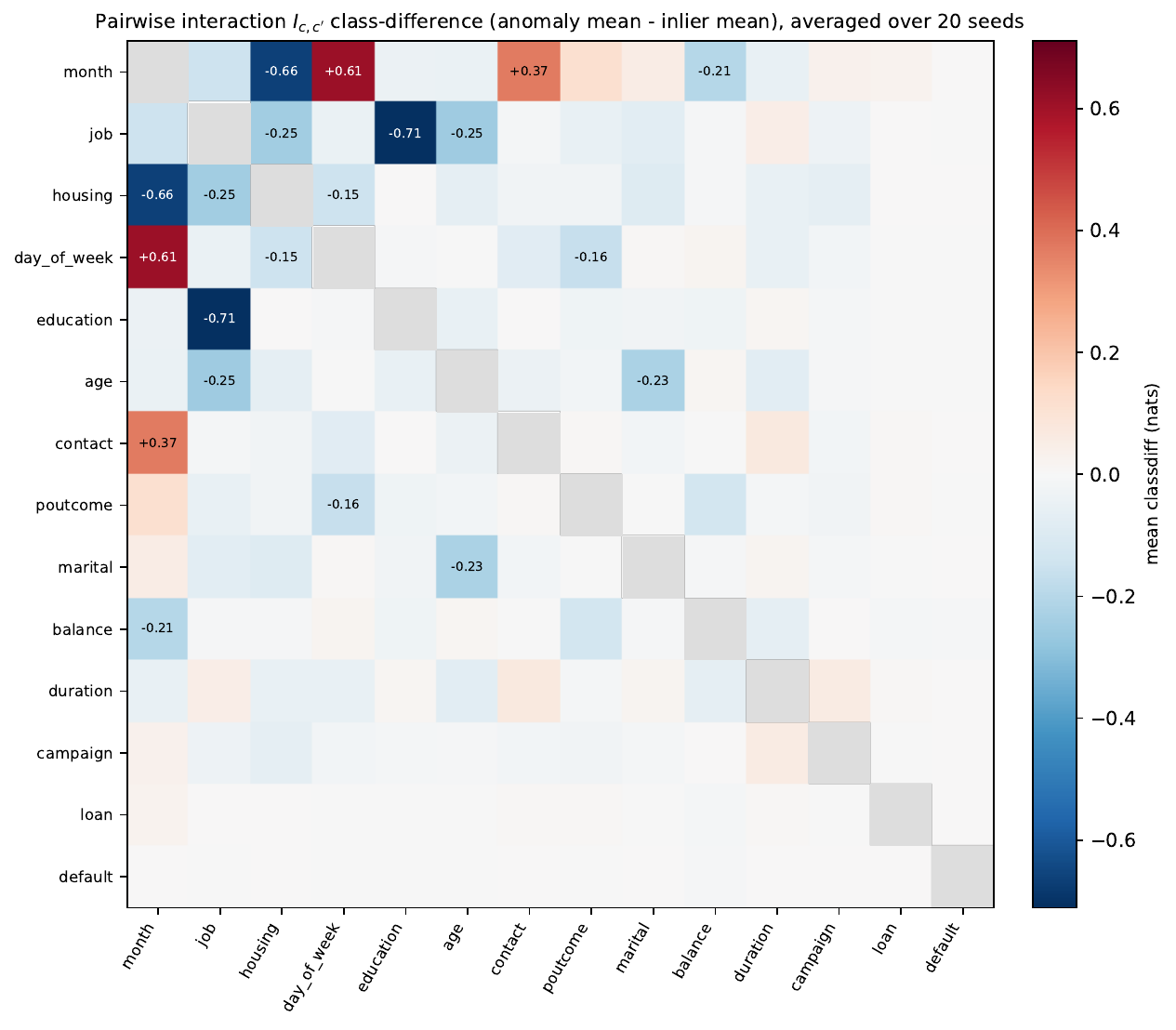}
\caption{Pairwise interaction effect $\bar{I}_{c, c'}$ across $n=20$ seeds. Red: pairs whose non-additive coupling contributes more to the anomaly score than to the inlier score; blue: the reverse. Bonferroni-significant cells ($p < 5.5{\times} 10^{-4}$) account for $31/91$ pairs.}
\label{fig:supp_pairwise_heatmap}
\end{figure}

Of the $\binom{14}{2} = 91$ pairs of effectively-active columns, $57/91$ are individually significant at $\alpha = 0.05$ uncorrected, and $31/91$ remain significant under Bonferroni correction ($\alpha / 91 \approx 5.5 \times 10^{-4}$) on a paired two-sided $t$-test of per-seed anomaly-vs-inlier means. Among those 31 pairs, the sign of $\bar{I}_{c, c'}$ is stable in every seed for the headline entries (top five pairs: $20/20$ seeds agree on sign), and across all 31 the fraction of seeds agreeing on sign is at least $0.9$. The remaining 60 pairs have small $|\bar{I}_{c, c'}|$ (median $0.02$ nats) and frequently flip sign across seeds, so the non-additive signal is concentrated in a small, robust subset rather than diffused across all pairs. Figure~\ref{fig:supp_pairwise_heatmap} visualises $\bar{I}_{c, c'}$ across all 91 pairs; Table~\ref{tab:supp_pairwise_top} lists the top five pairs by $|\bar{I}_{c, c'}|$.

\begin{table}[!ht]
\centering
\caption{Top five column pairs by $|\bar{I}_{c, c'}|$. ``frac.\ pos.'' = fraction of the 20 seeds for which the per-seed statistic is positive ($0.00$ / $1.00$ = full sign stability).}
\label{tab:supp_pairwise_top}
\begin{tabular}{llrrrl}
\toprule
$c$ & $c'$ & $\bar{I}_{c, c'}$ & SE & paired $t$ & frac.\ pos. \\
\midrule
job     & education    & $-0.711$ & $0.043$ & $-16.5$ & $0.00$ \\
housing & month        & $-0.664$ & $0.053$ & $-12.6$ & $0.00$ \\
month   & day\_of\_week & $+0.611$ & $0.048$ & $+12.6$ & $1.00$ \\
contact & month        & $+0.370$ & $0.026$ & $+14.4$ & $1.00$ \\
job     & age          & $-0.254$ & $0.011$ & $-23.8$ & $0.00$ \\
\bottomrule
\end{tabular}
\end{table}

The top pairs recover \emph{a priori} plausible joint structure: \texttt{job}\,$\times$\,\texttt{education} captures the occupational--educational correlation, \texttt{job}\,$\times$\,\texttt{age} and \texttt{marital}\,$\times$\,\texttt{age} capture demographic co-occurrence, and \texttt{month}\,$\times$\,$\{\texttt{day\_of\_week}, \texttt{contact}, \texttt{housing}\}$ captures the temporal--channel structure of a direct-marketing campaign. The per-pair $|\Delta F_{c, c'}|$ on its own reaches AUROC at most $0.71$ (\texttt{poutcome}\,$\times$\,\texttt{duration}), well below the headline $F_{\text{before}}$ score, so we frame $\bar{I}_{c, c'}$ as a structural interpretability diagnostic rather than a competing predictor.

\bibliographystyle{unsrtnat}
\bibliography{anomalyBM}

\end{document}